\documentclass[letterpaper, 10 pt, conference]{ieeeconf}  

\IEEEoverridecommandlockouts                              

\usepackage{todonotes}
\usepackage{amsmath}
\usepackage{algorithm2e}

\usepackage{algpseudocode}
\RestyleAlgo{ruled}
\usepackage{float}
\usepackage{amsfonts}

\title{\LARGE \bf A Scalable Reinforcement Learning Approach for Attack Allocation in Swarm to Swarm Engagement Problems
}

\author{Umut Demir$^{1}$ and Nazim Kemal Ure$^{2}$
\thanks{*This work was not supported by any organization}
\thanks{$^{1}$Umut Demir is with Istanbul Technical University, Artificial Intelligence and Data Science Application and Research Center, Istanbul,
Turkey.
        {\tt\small demiru@itu.edu.tr }}%
\thanks{$^{2}$Assoc. Prof. Dr. Nazim Kemal Ure with the Istanbul Technical University, Artificial Intelligence and Data Science Application and Research Center, Istanbul,
Turkey.
        {\tt\small ure@itu.edu.tr}}%
}

\begin{document}

\maketitle
\thispagestyle{empty}
\pagestyle{empty}

\begin{abstract}
In this work we propose a reinforcement learning (RL) framework that controls the density of a large-scale swarm for engaging with adversarial swarm attacks. Although there is a significant amount of existing work in applying artificial intelligence methods to swarm control, analysis of interactions between two adversarial swarms is a rather understudied area. Most of the existing work in this subject develop strategies by making hard assumptions regarding the strategy and dynamics of the adversarial swarm. Our main contribution is the formulation of the swarm to swarm engagement problem as a Markov Decision Process and development of RL algorithms that can compute engagement strategies without the knowledge of strategy/dynamics of the adversarial swarm. Simulation results show that the developed framework can handle a wide array of large-scale engagement scenarios in an efficient manner.
\end{abstract}

\section{INTRODUCTION}
Swarm robotics involve design of robust and scalable control and planning algorithms for coordinating a large numbers of robots. Collective behavior observed in nature provides inspiration to swarm robotics by showing how simple behaving individuals can achieve complex tasks when they gather in groups and cooperate. Recent works in artificial swarm intelligence and engineering can be found in extensive literature reviews of Brambilla \cite{brambilla2013swarm} and Schrabz \cite{schranz2020swarm}.

Aerial attack strategies have changed drastically in recent years due to advances in unmanned aerial vehicle technologies. The United States(DARPA OFFensive Swarm-Enabled Tactics), Russia(Lightning Project), France(Icarus Project) demonstrated that attacks can be carried out by a swarm of more than a hundred drones \cite{forbes}. DARPA's current OFFSET studies are focusing on swarms whose numbers exceed hundreds \cite{offset}. Since existing conventional aerial defense systems are optimized for a small number of heavy-hitting adversaries such as cruise missiles or fighter aircraft, these systems are often in a critical disadvantage against large-scale aerial swarm attacks that cover a wide area. Thus, defending against the aerial swarm attacks is one of the most important issues in the modern defense industry~\cite{williams2018swarm}.

With the advances in reinforcement learning \cite{silver2016mastering} and swarm robotics, the merging of these two areas has become inevitable for intelligent swarm systems that does not rely on heuristics and self-learn attack and defense strategies from past interactions. Batra et al. \cite{batra2022decentralized} have developed drone swarm planning algorithms for a large number of drones using end-to-end reinforcement learning in a decentralized manner. Bredeche and Fontbonne \cite{bredeche2022social} discussed distributed online reinforcement learning methods to achieve social swarm intelligence.

In addition to control algorithms, Hansen and Brunton \cite{hansen2022swarm} have proposed SwarmDMD to learn and reconstruct swarm dynamics. They use data-driven method combined with dynamic mode decomposition to learn local agent interactions and swarm behaviour. This study has also been a source of inspiration for the model proposed in this work. Furthermore, many social sciences and biology studies have been conducted on nature inspired swarm systems. Well-known study of Viscek \cite{vicsek1995novel} showed that flocking is a kind of self alignment and very simple local interactions can lead to complex swarm behaviours. This work is also used in SwarmDMD \cite{hansen2022swarm} to learn swarm dynamics as a local interaction model, which is also used in our proposed work. Many variations of these studies can be found in the literature but these subjects are beyond the scope of this paper.


Swarm-to-Swarm engagement problem focuses on computing optimal strategies of controlled swarm units against an adversarial swarm \cite{uzun2020probabilistic}. Although there exists previous work on these subjects, there are no efficient and scalable methods for solving swarm to swarm engagement problems without using strict assumptions and constraints. The previous works show scalable
results for the high level engagement problem using both deterministic \cite{eren2017velocity}, \cite{zhao2011density} and probabilistic methods \cite{uzun2020probabilistic}, \cite{demir2015decentralized}, \cite{uzun2020decentralized} by ignoring engagement scenarios where the adversarial swarm adapts to controlled swarm actions and to the changes in the environment. For instance, in \cite{uzun2020probabilistic} it is assumed that adversarial swarm directly fly towards to a predefined destination. Although this approach is very useful for optimality analysis, its applicability to real scenarios is limited. One of the central objectives in this papers is to develop algorithms that are applicable to a wider array of swarm to swarm engagement problems without using hard assumptions regarding the nature of the adversarial swarm.


\subsection{Contributions}
The main contribution of this work is a reinforcement learning (RL) framework that computes engagement plans against large-scale swarm to swarm attacks, without relying on any hard assumptions regarding the strategy and dynamics of the adversarial swarm. In particular:

\begin{itemize}
\item Based on the work in~\cite{hansen2022swarm}, we formulate the swarm to swarm engagement problem as a Markov Decision Process (MDP), which enables using RL algorithms for computing engagement policies without the knowledge of the underlying dynamics of the adversarial swarm. 

\item Simulation results show that the proposed framework is capable of computing efficient engagement strategies for a variety of scenarios under different initial conditions and number of swarm units.
\end{itemize}


\section{PRELIMINARIES}

In this section, background material and concepts used in the development of the framework are presented. 
\begin{figure*}[t!]
    \centering
    \includegraphics[width=0.95\textwidth]{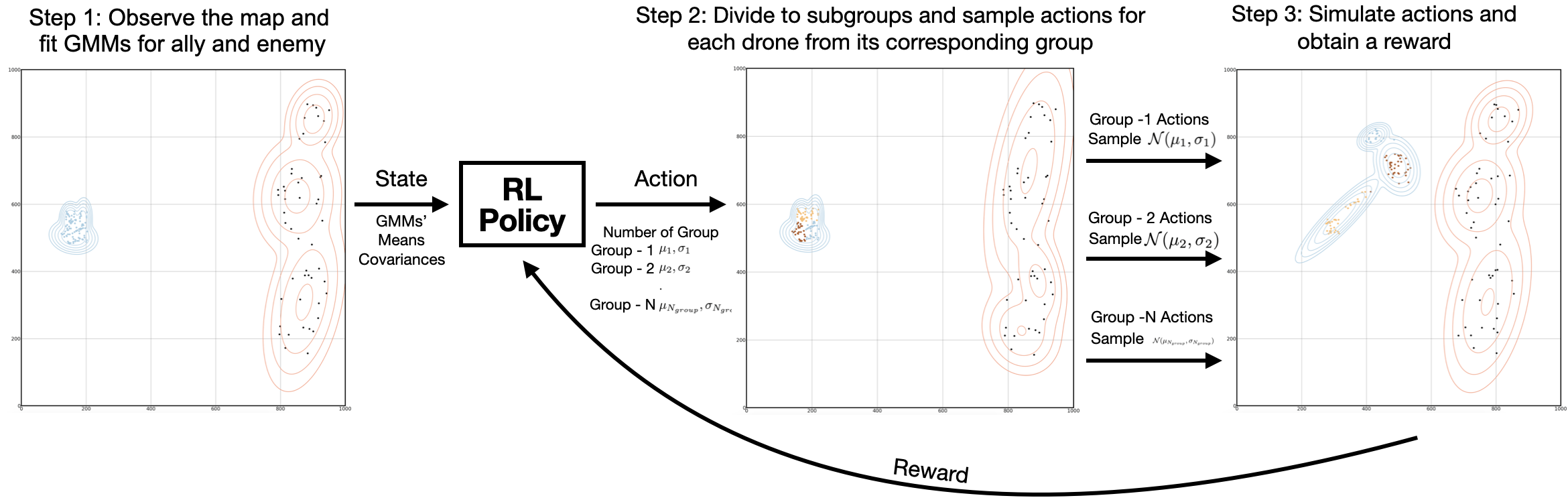}
    \caption{The proposed swarm-to-swarm learning and planning framework. In the first figure, the blue and black dots are the controlled agents and adversarial agents, respectively. The blue lines and red curves are the Gaussian mixture models. In the second figure, the RL policy controls how many groups to divide the swarm into and generates control inputs for each group. In the last figure, agents belonging to the group samples action from the group distribution and simulates these actions. Finally, a reward signal is fed back to RL algorithm.}
    \label{fig:diagram}
\end{figure*}
\subsection{Swarm Unit Dynamics} \label{dynamicssection}
In this section, swarm unit dynamics are explained. Although simplified dynamics are considered here for high-level planning, the dynamics of the system can be replaced with the dynamics of desired fidelity.

We consider a two dimensional (2-D) map with continuous coordinates. Agent-i's state vector, $\boldsymbol{x^i}$, consists of its position, heading angle, velocity and angular velocity at the discrete time $t$, for $i \in (1,2,3,...N_{agent}) $ where $N_{agent}$ is the number of agents. The state of the swarm,  $\boldsymbol{X_t}$, consists of states of every agent at time, $t$,
\begin{align}
    \boldsymbol{x_t^i} &= [x_t^i, y_t^i, \theta_t^i, v_t^i, \omega_t^i]^T \\
    \boldsymbol{X_t} &= [\boldsymbol{x^1_t}, \boldsymbol{x^2_t}, ..., \boldsymbol{x^{N_{agent}}_t]^T }.
    \label{state}
\end{align}
For small $\Delta t $, the control input for agent-i $\boldsymbol{u^i}$ can be defined as;
\begin{align}
    \boldsymbol{u^i} = [u_v^i, u_w^i]^T.
    \label{controlu}
\end{align}
After $\Delta t $ amount of time passes, change of agent-i state can be calculated as in Equation \ref{transition}: \cite{thrun2002probabilistic}.
\begin{align}
\label{transition}
        \begin{pmatrix} x^i_t \\ y^i_t \\ \theta^i_t \\ v^i_t \\ \omega^i_t \end{pmatrix} = \begin{pmatrix} x^i_{t-1} -\frac{v^i_t}{\omega^i_t}sin(\theta^i_{t-1}) + \frac{v^i_t}{\omega^i_t}sin(\theta^i_{t-1} + \omega^i_t \Delta t ) \\ 
        y^i_{t-1} + \frac{v^i_t}{\omega^i_t}cos(\theta^i_{t-1}) - \frac{v^i_t}{\omega^i_t}cos(\theta^i_{t-1} + \omega^i_t \Delta t ) \\
        \theta^i_{t-1} + \omega^i_t \Delta t \\
        v^i_{t-1} + u^i_{v,t} \Delta t \\
        \omega^i_{t-1} + u^i_{w,t} \Delta t
         \end{pmatrix}.
\end{align}
In reality, despite a direct transition from one velocity to another is not possible, the change in velocity is controlled by adding a limit and keeping time step as small as possible.

\subsection{Swarm Dynamics and Flocking}
In this section, the considered swarm behaviour and flocking are explained.

We take a similar approach to the Viscek model. The Vicsek model is a simple model used for describing collective motion and swarming \cite{vicsek1995novel}. In the original Vicsek model, agents maintains a constant forward speed and interact with the swarm by aligning their heading direction with the average heading of their neighbours in a certain Euclidean radius. In other words, it assumes that flocking is a kind of self alignment behaviour.

According to the model, position vector of agent-i at time $t$ $\boldsymbol{r_i(t)}$ and direction of heading angle of the velocity $\boldsymbol{\theta_i(t)}$ follows following rules;
\begin{enumerate}
    \item Agents align itself with their neighbours within a euclidean distance $r$ with a noise $\eta_i(t)$
    \begin{itemize}
        \item $\boldsymbol{\theta_i(t+\Delta t)} = \langle \theta_j \rangle	_{ \mid r_i - r_j \mid < r} + \eta_i(t) $
    \end{itemize}
    \item Then agent continuous to follow constant speed in the aligned direction
    \begin{itemize}
        \item $ \boldsymbol{r_i(t+\Delta t)} =\boldsymbol{r_i(t)} + v \Delta t  \begin{pmatrix} cos\theta_i(t) \\  sin \theta_i(t) \end{pmatrix} $
    \end{itemize}

\end{enumerate}
Based on Viscek's approach, controlling the mean behavior of the swarm and forcing swarm agents follow the mean behaviour of the swarm leads to a scalable control framework.

First, to control mean behaviour of the swarm, instead of using Eq. \ref{controlu} for individual agents, we generate mean control input of the swarm. This input is the mean value of the control input to be applied to each unit throughout the swarm.

Secondly, we also control a variance value. Therefore, the control system acts like a distribution with a mean and a variance value. To distribute control inputs to each swarm units, every unit samples from this distribution to obtain its individual control inputs.

The control input for the swarm can be seen at Eq. \ref{controlu_mean}, \ref{controlu_var}.
\begin{align}
    \boldsymbol{u_{\mu_{swarm}}} = [u_{v, \mu_{swarm}}, u_{\omega, \mu_{swarm}}]^T \label{controlu_mean} \\     
 \boldsymbol{u_{\sigma^2_{swarm}}} = [u_{v, \sigma^2_{swarm}}, u_{\omega, \sigma^2_{swarm}}]^T \label{controlu_var}
\end{align}

Each agent belonging to swarm samples a control input from a normal distribution, $ \mathcal{N} (u_{\mu_{swarm}}, u_{\sigma^2_{swarm}})$ with a probability density function (PDF) at Eq. \ref{propdist}.
\begin{align}
        \label{propdist} &\boldsymbol{p(x)} = \frac{1}{\sqrt{2 \pi \boldsymbol{\sigma^2}}} e^{- \frac{(x- \boldsymbol{\mu})^2}{2 \boldsymbol{\sigma^2}}} \\
    \notag &\text{where } \boldsymbol{\mu} = \boldsymbol{ u_{\mu_{swarm}} } \text{ and } \boldsymbol{\sigma^2} = \boldsymbol{ u_{\sigma^2_{swarm}} }.
\end{align}
 The goal here is to control the distribution of the swarm by controlling the mean and variance values. The importance of controlling the variance value here is to control how much the swarm is drawn to the mean value. A low variance value decreases the spread of the swarm on the map and brings the swarm agents closer together, while a large variance value increases the spread on the map and distances the swarm agents. Thus, we obtain control over the swarm distribution by controlling only mean and variance parameters.

\subsection{Swarm Groups}
Controlling the whole swarm with a single control input reduces the maneuvering capability of the swarm. To increase capability while keeping complexity low, we decompose the swarm into sub-swarm groups.

The decision of how many groups to divide the swarm into is decided by the RL policy. The goal here is to reduce the complexity by keeping the number of groups as low as possible, but also to benefit from grouping.

We expect to learn intelligent grouping decisions from the RL policy. The number of groups decided by policy as in Equation \ref{grouping}.
\begin{align}
\label{grouping}
     \{ N_{group} \mid 1 \leq N_{group} \leq  N_{group, max}, N_{group} \in \mathbb{Z} \}
\end{align}
where $N_{group}$ is the number of groups to divide, $N_{group, max}$ is maximum possible number of groups.

The k-means algorithm is used for grouping drones that are close to each other and express similar behavior in the same group. To find which swarm unit belongs to which swarm group, k-means clustering algorithm is executed every decision making step. 

\subsection{Swarm Observation}
In this section, use of Gaussian Mixture Models as representation of the swarm distribution is explained.

Using the state space of the swarm at Eq. \ref{state} as observation is not feasible for two reasons. 
\begin{enumerate}
    \item The first motivation is that size of the state space increases with the number of agents. This not scalable and problem becomes unmanageable when the number of agents exceeds hundreds. 
    \item The second motivation is that the change of the observation space size is not desirable for neural networks architectures.
\end{enumerate}

Therefore, it is useful to approach to swarm as a distribution on a map and use the mean and covariance of the distributions. The Gaussian mixture model is preferred to represent the distribution on the map.

\begin{algorithm}[]
\caption{Allocation Algorithm}
\label{alg:algorithm} 
\begin{enumerate}
    \item Observe states and obtain $\boldsymbol{X_{controlled}}$ $\boldsymbol{X_{adversarial}}$ using Equation \ref{state}.
    \item Fit GMMs and obtain State $\boldsymbol{S}$ using Equation \ref{rlstate}.
    \item Obtain action from the policy $ \boldsymbol{A} \sim \pi(\boldsymbol{S}) $ 
    \item Fit K-Means Clustering with $N_{group}$ cluster centers. Assign swarm units to its groups according to K-means scores.
    \item Sample control input for every agent from its group distribution $\boldsymbol{u^i}  \sim \mathcal{N} (u_{\mu_{group}}, u_{\sigma^2_{group}}) $ using Equation \ref{propdist}
    \item Simulate actions and store $(s,a,r,s',d)$ on replay buffer for training
\end{enumerate}
\end{algorithm}

\subsection{Communication, Command and Control}
High capacity communication network is required for a large number of unmanned aerial vehicles to operate in cooperation. 
\begin{figure}[]
    \centering
    \includegraphics[width=\columnwidth]{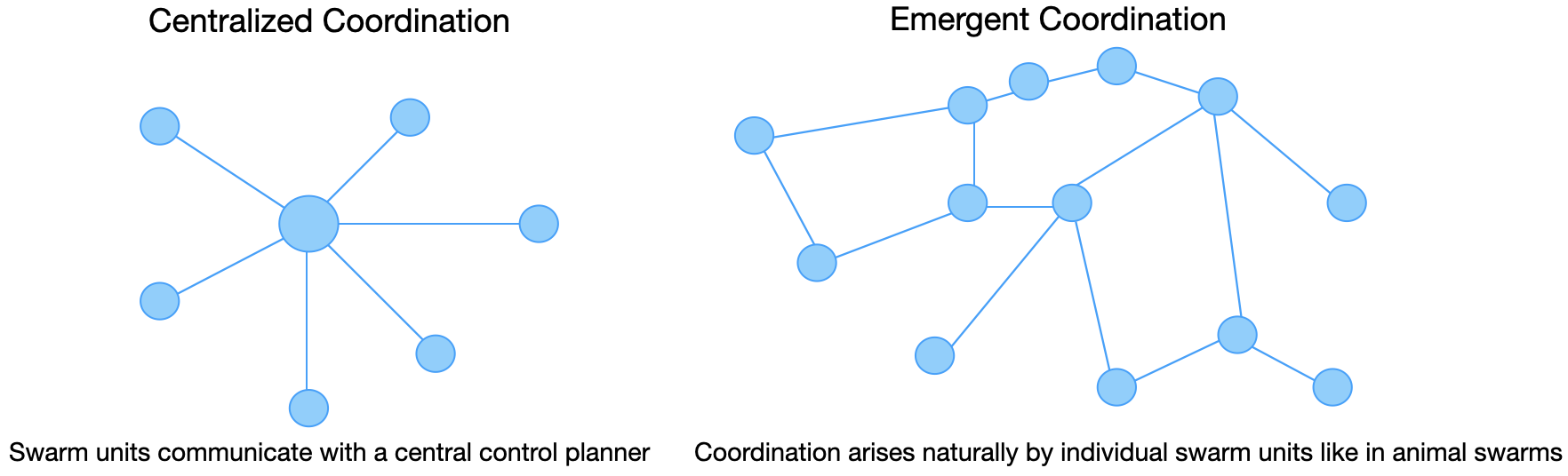}
    \caption{Swarm Command and Control Models}
    \label{fig:reward}
\end{figure}
Each of these models has different advantages, however, communication of all agents with each other in very large areas is not a good option for this problem. Control by a single central agent both creates communication problems and causes loss of control when central drone fails \cite{williams2018swarm}. Adapting changing circumstances such as using bandwidth when it is available or to decentralized decision-making when it is not, is the best option for this problem \cite{scharre2014robotics}.

Our framework uses an approximation of the distribution via GMMs as the state space. For this reason, there is no need for constant communication with every swarm unit. Swarm units do not know what the other swarm units on the other side of the swarm are doing. Due to its adaptive nature, Emergent Coordination is reliable and best fit to our framework.

\section{Reinforcement Learning for Swarms Allocation}

In this section, our approach to the problem within the framework of reinforcement learning is explained.
\subsection{Markov Decision Process}
We consider a Markov Decision Process~(MDP) \cite{sutton2018reinforcement} with discrete time steps where a single agent observes full map state and generates continuous actions until termination conditions are satisfied. The MDP consists of 5 parameters $\langle S,A,T,R,\gamma \rangle$, where  $S$ is the state space, A is the action space, $T(S_{t+1}|S_t,A_t)$ is the transition dynamics of the agents stated at Section \ref{dynamicssection} and $\gamma \in [0,1)$ is the discount factor and $R$ is the reward obtained from the transition. The agent actions are determined with a policy function $\pi (a \mid s; \theta)$ parameterized by $\theta$ that outputs continuous actions.
The main objective of the problem is to maximize the discounted cumulative reward,
\begin{equation}
    \label{eq:exp}
    \displaystyle \mathop{\mathbb{E}}_{\pi, \gamma} \left[\sum_{i=0}^{T} \gamma^{i} R_i \mid S_0 = s \right]
\end{equation} 

The state space of the Markov Decision Process consists of Gaussian Mixture Model parameters of the position distribution of the swarms. Every decision step swarm distribution is fitted to a Gaussian Mixture Model with predetermined number of clusters, $ N_{cluster} $.

\begin{align}
& \boldsymbol{\mu_{n}} = [ \mu_{n,x}, \mu_{n,y} ]_{1x2}, \sigma_{n} = \begin{pmatrix} \sigma_{n,x,x} & \sigma_{n,x,y} \\
    \sigma_{n,y,x} & \sigma_{n,y,y} \end{pmatrix}_{2x2} \\
    &\text{for } n = 1 \text{ to } N_{cluster} \notag
\end{align}
In the flat vector form, the covariance and the state vector can be written as follows;
\begin{align}
    \boldsymbol{\sigma_{n, f}} &= [ \sigma_{n,x,x}, \sigma_{n,x,y}, \sigma_{n,y,x}, \sigma_{n,y,y} ]_{1x4} \\
\boldsymbol{S_{swarm}} &= [ \mu_n , \sigma_{n,f}]_{1x6N_{cluster}} \text{ for } n = 1 \text{ to } N_{cluster} \\
\boldsymbol{S} &= [\boldsymbol{S_{controlled}}, \boldsymbol{S_{adversarial}}]
\label{rlstate}
\end{align}

The action space of the MDP consists of number of groups, means and variances of the groups.
\begin{align}
    \boldsymbol{A} = [ N_{group}, \textbf{u}^n_{\mu}, \textbf{u}^n_{\sigma^2}] \text{ for } n = 1 \text{ to } N_{group, max}
\end{align}
where $\textbf{u}^n_{\mu}, \textbf{u}^n_{\sigma^2}$ are means and variances of the control inputs of group-n. Here, although the maximum number of control inputs are generated at each step, only as many control inputs as the number of groups are used. The remaining control inputs are not used.

\subsection{Reward}
Here, two different reward structures for two different scenarios are used.
\begin{enumerate}
    \item Reward - 1: Only the elimination of the adversarial swarm is addressed.
    \item Reward - 2: In addition to eliminating the adversarial swarm, it is also taken into account to gain quantitative superiority. If an adversarial unit falls within the radius of impact of more than one controlled unit, policy gains an additional reward.
\end{enumerate}
The Reward-1 aims to eliminate adversarial units as fast as possible. On the other hand, the Reward-2 aims to eliminate the adversarial units by quantitative superiority.
\subsection{Policy and Training}
This section describes the training of policy using reinforcement learning.

In this work, Twin Delayed DDPG(TD3) algorithm, an improved version of DDPG, to learn a policy is used \cite{fujimoto2018addressing} \cite{TD3}. There are many reinforcement learning algorithms that have proven their success in dynamic systems. Deep Deterministic Policy Gradient (DDPG) is an algorithm that simultaneously learns a Q function and a policy. It is an off-policy algorithm which uses the Bellman equation to learn the Q function and utilizes the Q function to learn the policy. However, DDPG algorithm mostly overestimates Q-values which result with undesired policies. Using clipped q-learning, delayed policy updates and target-policy smoothing, TD3 algorithm results with more stable and better policies. The TD3 algorithm can be seen at Algorithm \ref{alg:TD3}.

The policy consist of a features extractor followed by a fully-connected network. The features extractor whose role is to extract features from high-dimensional space to lower dimension space. The fully-connected network that maps the feature extractor outputs to actions and q-values. The extractor have 1024 neurons. Actor critic networks have 3 layers with 500, 300 and 100 neurons, respectively. Simple experiments show that the LSTM based networks rise better results however, very close and much faster results are achievable with simple fully connected networks.

\begin{algorithm}[]
\caption{Twin Delayed DDPG \cite{fujimoto2018addressing} \cite{TD3}} \label{alg:TD3} 
Input: Initial Policy Parameters $\theta$, Q-function parameters $\phi_1, \phi_2$, empty replay buffer $D$ \\
Set target parameters to initial parameters $\theta_{target} \gets \theta$, $\phi_{target, 1} \gets \phi_1 $, $\phi_{target,2} \gets \phi_2$ \\
\textbf{Repeat} \\
Select an action: $a = \mu_{\theta(s)} + \epsilon $, where $s$ is state, $ \epsilon \sim \mathcal{N} $ and $ a \in [a_{low}, a_{high}]$ \\
Execute action $a$ and obtain next state, $s'$ \\
Store $(s,a,r,s',d)$ replay buffer, $D$ \\
If $s'$ is terminal, reset the environment \\
\If{it's time to update}{
        \For{j in range} {Randomly sample a batch of transitions, $B=(s,a,r,s',d)$ from $D$ \\
        Compute target actions $a'(s') = \mu_{\theta(s')} + clip( \epsilon, -c, c)  $ \\
        Compute targets \\ $y(r, s', d) = r + \gamma(1-d)\underset{i=1,2}{\text{min}} Q_{\phi_{target,i}}(s', a'(s'))$ \\
        \If{$j \text{ mod } \text{policy delay} == 0$}{
        Update Q-function with gradient descent using \\ $\nabla_{\theta_i}\frac{1}{\mid B \mid} \sum_{(s,a,r,s',d) \in B}  ( Q_{\phi_i}(s,a) - y(r,s',d))^2 $ \\
        Update target networks with \\
        $\phi_{target,i} \gets \rho \phi_{target,i} + (1-\rho)\phi_{i} $ \\
        $\theta_{target,i} \gets \rho \theta_{target,i} + (1-\rho)\theta_{i} $ } } }
\textbf{Until}{Converge}
\end{algorithm}

\section{Simulation Results}
In this section, the simulation conditions of the experiments are given and the results are discussed.

We present results show that different defense mechanisms develop according to different reward collection mechanism and enemy strategies. To give an example, the enemy can attack spread over the map, consecutively, or clustered in a single center. All of these attacks require separate defense mechanisms. On the other hand, if defending as a group of swarm provides a numerical advantage, this will result in a different defense mechanism brought by a different reward structure. The main objective is to see a control policy developed that adapts to these different scenarios.

\subsection{Scenarios}
The simulation conditions for the controlled swarms are presented at Table \ref{tab:simcon}.

Each swarm element's state vector is sampled from a normal distribution with predefined parameters according to their scenarios.

\begin{table}[H]
\caption{Simulation Conditions}
\label{tab:simcon}
\resizebox{\columnwidth}{!}{%
\begin{tabular}{|l|c|l|c|}
\hline
Simulation Timestep & $0.1 s$  & RL Timestep       & $1 s$     \\ \hline
$v_{min}$           & $30 m/s$ & $v_{max}$         & $300 m/s$ \\ \hline
$w_{min}$           & $-\pi/5$  & $w_{max}$         & $-\pi/5$   \\ \hline
Impact Distance     & $30 m$   & $N_{group, max}$ & $5$       \\ \hline
$N_{cluster}$     & $3$   & GMM and K-means Accuracy & $0.01$       \\ \hline
\end{tabular}%
}
\end{table}

\begin{figure*}[t!]
    \centering
    \includegraphics[width=\textwidth]{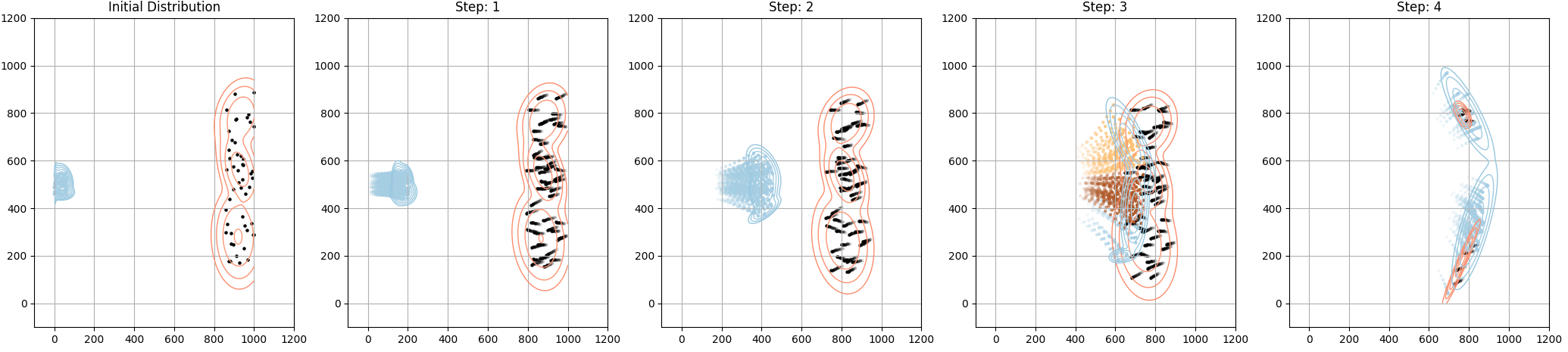}
    \caption{Simulation Results of Scenario - 1: 100 Controlled Swarm Units(Blue Dots), 50 Adversarial Swarm Units(Black Dots), All adversarial units are eliminated at end of Step-4}
    \label{fig:result1}
\end{figure*}

\begin{figure*}[t!]
    \centering
    \includegraphics[width=\textwidth]{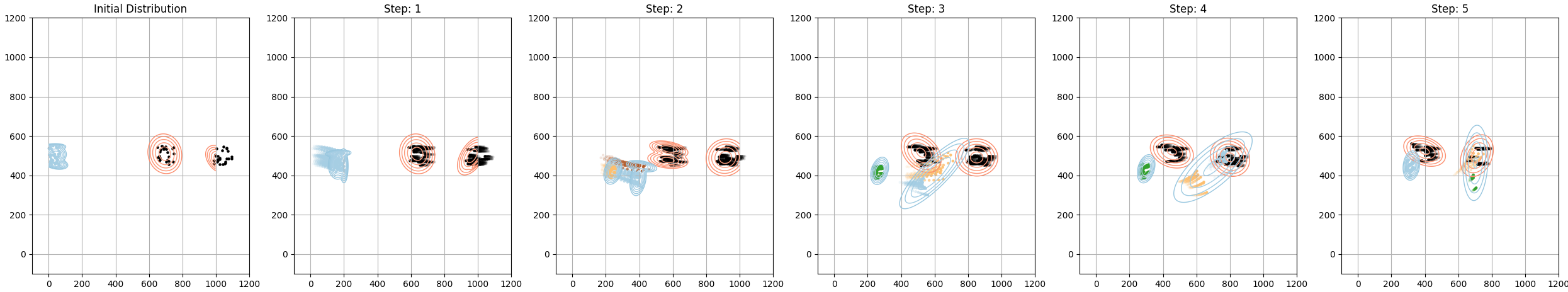}
    \caption{Simulation Results of Scenario - 2: 50 Controlled Swarm Units(Blue Dots), 50 Adversarial Swarm Units(Black Dots), All adversarial units are eliminated at end of Step-5}
    \label{fig:result2}
\end{figure*}
\begin{figure*}[t!]
    \centering
    \includegraphics[width=\textwidth]{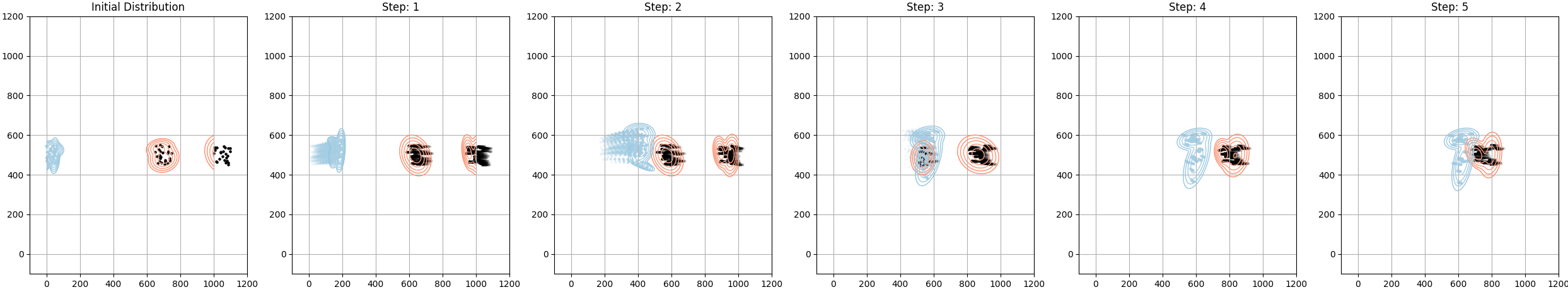}
    \caption{Simulation Results of Scenario - 3: 50 Controlled Swarm Units(Blue Dots), 50 Adversarial Swarm Units(Black Dots),  All adversarial units are eliminated at end of Step-5}
    \label{fig:result3}
\end{figure*}

\subsubsection{Scenario - 1}
The purpose of the first scenario is to show how to change the distribution by taking advantage of the flexibility of grouping and controlling the variance.

Initially, the controlled swarm is concentrated in one small area. The adversarial swarm comes from the right side and covers more area on the map as seen at Figure \ref{fig:result1}. Using the reward function Reward-1 it is expected from policy to eliminate adversarial units as soon as possible.

In the Step-1 and Step-2 of Figure  \ref{fig:result1}, it is seen that the controlled swarm units accelerate towards adversarial swarm by controlling the mean value. In addition, it is seen that it increases its distribution on the map by controlling the variance of the swarm. Finally, at Step-3 the policy divides the swarm into three groups and directs each swarm group towards the centers of the adversarial swarm.

The results show that policy utilizes grouping decisions, mean value control and standard deviation control to achieve maximum possible reward.

\subsubsection{Scenario - 2}
In the second scenario adversarial swarms move to the left side into two swarm groups as can be seen at Figure \ref{fig:result2}.

Similar to the Scenario-1, the reward function Reward-1 is used and a quick elimination of adversarial units is expected.

In the Step-1 of Figure \ref{fig:result2}, it can be seen that controlled swarm accelerates to the enemy units. Further, at Step-3 units are divided to three groups and two of them further accelerate to reach second wave of the adversarial units. As green group interacts with the front part of the swarm, two fast groups attack the swarm in the back.

The results show that policy utilizes grouping decisions and mean value control to achieve quick elimination of the adversarial units. Unlike to Scenario-1, there is no variance control as expected since all adversarial units are clustered in small area.

\subsubsection{Scenario - 3}
The simulation conditions discussed in this scenario are exactly the same with Scenario-2 except for the reward function. The reward function Reward-2 is used to show that policy learns to outnumber adversarial units. 

In the Figure \ref{fig:result3}, it can be  seen that, unlike to Step-2 of Figure \ref{fig:result2}, the controlled swarm agents stay as group and obtain numerical advantage against first attack wave of the adversarial swarm. After the elimination of the first wave at the end of the Step-3, the remaining controlled swarm units allocate the second wave.

This result is expected since Reward-2 is also taken into account to gain quantitative superiority.

\subsection{Training Different Scenarios with Curriculum Learning}
Curriculum learning is a type of learning where learning starts with easy examples of a task and then gradually increase the task difficulty \cite{bengio2009curriculum}.

In the experiments, it was observed that giving all the scenarios at the same time reduced the performance. By using curricular learning, the problem was transformed into a learning structure, starting with the easy ones and increasing the difficulty in order.

The easiest scenario was chosen as the scenario where the adversarial distribution is in the center of the map and the the controlled distribution is similar to this distribution initially. Further, to give advantage to the controlled distribution, a high number of agents has been given. After a successfully training on simple scenario, Scenario-1, Scenario-2, Scenario-3 are added, respectively. While doing this, the number of the controlled agents has been reduced towards the number of adversarial units.

\begin{figure}[]
    \centering
    \includegraphics[width=\columnwidth]{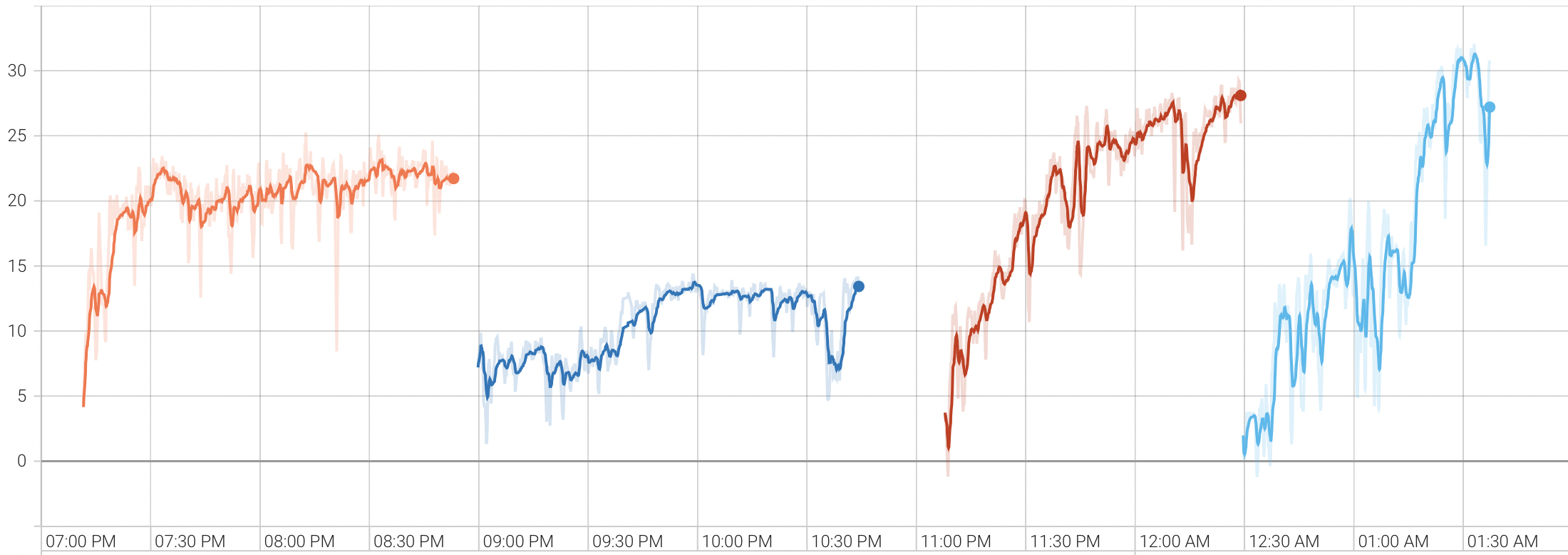}
    \caption{Reward and Training: Total Environment Steps is 800k in 6 Hours, Color change represents the addition of a new scenario, Maximum Environment Reward is 35}
    \label{fig:reward1}
\end{figure}

The amount of reward collected during the training is shown in Figure \ref{fig:reward1}. Each color change represents the addition of a new scenario.

\subsection{Simulation Time}
All tests were done on Macbook Pro M1 with 16GB RAM.
Training for 4 scenarios takes 6 hours for 800000 time steps and converges to maximum possible environment reward. Averaged over 1000 simulation for 100 drones, single simulation step without training is 0.025 seconds. 

\section{Conclusion}
In this study, reinforcement learning
framework that controls density of a large-scale, controlled swarm in continuous environment to engage adversarial swarm attacks are developed. Inspired by the simple interactions of swarm units seen in nature, we found a simple framework where swarm elements follow the mean behaviour of the swarm. Further, we found increasing the flexibility of this behavior in the grouping of the swarm that controlled by policy.

The results show that this framework adapts to the most common swarm attack methods. On the other hand, it needs to be tested in more scenarios where enemy acts intelligently. In order to achieve this, a framework in which policies compete with each other as in AlphaGo \cite{silver2016mastering} would give result in interesting strategies.

In addition, in order to achieve realistic combat simulations, as we did in the previous study \cite{demir2022}, the differential pursuit-evasion games can be run when controlled and adversarial units engage.

\newpage
\bibliographystyle{IEEEtran}
\bibliography{bibi}

\end{document}